# Efficient MRF Energy Minimization via Adaptive Diminishing Smoothing

**Bogdan Savchynskyy**[1]    **Stefan Schmidt**[1]    **Jörg Kappes**[2]    **Christoph Schnörr**[1,2]
[1]Heidelberg Collaboratory for Image Processing, Heidelberg University, Germany,
[2]Image and Pattern Analysis Group, Heidelberg University, Germany

## Abstract

We consider the linear programming relaxation of an energy minimization problem for Markov Random Fields. The dual objective of this problem can be treated as a concave and unconstrained, but non-smooth function. The idea of smoothing the objective prior to optimization was recently proposed in a series of papers. Some of them suggested the idea to decrease the amount of smoothing (so called temperature) while getting closer to the optimum. However, no theoretical substantiation was provided.

We propose an adaptive smoothing diminishing algorithm based on the duality gap between relaxed primal and dual objectives and demonstrate the efficiency of our approach with a smoothed version of Sequential Tree-Reweighted Message Passing (TRW-S) algorithm. The strategy is applicable to other algorithms as well, avoids ad-hoc tuning of the smoothing during iterations, and provably guarantees convergence to the optimum.

## 1 INTRODUCTION

We consider the problem of computing the most likely configuration $x$ for a given graphical model, i.e. a distribution $p_\mathcal{G}(x;\theta) \propto \exp(-E_\mathcal{G}(\theta,x))$. The problem to compute the most likely labeling $x$ (*MAP labeling problem*) amounts to minimizing the energy function[1]

$$\min_{x\in\mathcal{X}} E_\mathcal{G}(\theta,x) = \min_{x\in\mathcal{X}} \left\{ \sum_{v\in\mathcal{V}} \theta_{v,x_v} + \sum_{uv\in\mathcal{E}} \theta_{uv,x_{uv}} \right\}. \quad (1)$$

While problem (1) is known to be NP-hard, we will concentrate mainly on the linear programming (LP) relaxation of the problem, originally proposed by Schlesinger [16] – see [24] for a recent review.

A popular approach for solving the relaxed problem is based on maximization of its dual objective, which constitutes a lower bound for the initial objective (1). It is well-known that the dual can be treated as a concave but non-smooth unconstrained problem (see e.g. [17]). There are a number of algorithmic schemes targeting it in its original non-smooth form, e.g. [16, 24, 8, 17, 9, 4]. Some of them, namely sub-gradient algorithms [17, 9] are guaranteed to reach its solution, but are extremely slow, others – message passing (e.g. [8, 4, 11]) – typically perform faster, but may get stuck in non-optimal points, since they can be considered as (block-)coordinate ascent.

Slow convergence of sub-gradient methods and stalling of message-passing ones are caused by the non-smoothness of the objective. Hence the idea of applying smoothing was proposed in a series of recent papers [6, 25, 7, 15]. However to reach a good approximate solution of the initial non-smooth objective the smoothing degree should be selected properly. There are two approaches how this can be done:
(i) **Precision oriented smoothing** approach [7, 15], following the fundamental paper [12]. The smoothing degree depends on the precision to be achieved and is selected and held fixed (or it changes only slightly) during the algorithm iterations.
(ii) Iterative or **diminishing smoothing** approach follows the idea to decrease the smoothing degree as the current iterate gets closer to the optimum. So far we are not aware of any algorithm employing this scenario for the problem (1), although the basic idea is mentioned in several papers, cf. [6, 25].

The idea of the diminishing smoothing corresponds to the following intuition: as long the iterate is far from the optimum one can use a coarse (very smooth) approximation of the objective, since it allows for faster optimization. The closer the iterate is to the optimum, the finer (and thus less smooth and computationally more costly) the approximation is required to guarantee a certain solution accuracy.

---
[1]We describe our notation in Section 2.

Both, precision oriented and diminishing smoothing approaches operate with the *smoothing gap* i.e. the largest difference between an objective function and its smooth approximation. The smoothing gap however can be controlled only indirectly via some *smoothing parameter*. In its turn the smoothing parameter can be set based on a **worst-case** analysis describing its influence on the smoothing gap. Alternatively this influence can be estimated **adaptively**, as discussed in [15].

For our diminishing smoothing approach some method for estimating the upper bound for the LP relaxation of the energy minimization problem is required. There are several recent papers addressing this issue [26, 18, 15] and we found that the method proposed in [15] fits our needs best. Our algorithm itself was inspired by [13], but in contrast we consider a purely dual optimization instead of the primal-dual one (to avoid massive memory allocation for the primal problem). Furthermore we focus on efficiency of the algorithm on average rather than in the worst-case by exploiting the specific structure of our problem.

We demonstrate the efficiency of our approach with a smooth version of the Sequential Tree-Reweighted Message Passing (S-TRWS) algorithm evaluated on the Middlebury database [20] and a series of computer generated examples. This algorithm, proposed for fixed smoothing in [11] and partially analyzed in [25] estimates the tree-reweighted free energy and can be considered as a smooth version of TRW-S [8]. Contrary to the fixed smoothing we consider a diminishing sequence of the smoothing parameter and show that this solves the LP relaxation of the energy minimization problem (1) up to a given precision. We compare our approach to the S-TRWS algorithm with a final precision oriented smoothing and a Nesterov first-order smoothing based optimization scheme proposed in [7] and [15] and show that it significantly outperforms all of them.

Without loss of generality, we concentrate on the common special case of grid-structured models here (as the benchmark [20] deals with this setting), but our results apply to non-grid-structured graphs as well.

**Contribution.** The contribution of this paper is four-fold:

1. We describe a diminishing smoothing algorithm for the LP relaxation of the problem (1), based on the duality gap of the relaxed problem.

2. We analyze the method of adaptive selection of the smoothing parameter proposed in [15] and show that in general it does not guarantee attainment of the prescribed precision (as implicitly stated in [15]). We propose a modification of the method, which eliminates this disadvantage, while preserves its efficiency.

3. Additionally, we propose a method for obtaining the sound stopping criterion for algorithms which maximize the *tree-reweighted free energy* [22] via minimizing its dual [5, 3, 11]. To this end we provide a method for computing a primal bound from a current dual iterate. This bound approaches the optimum of the *tree-reweighted free energy* together with the dual iterates.

4. We evaluate the S-TRWS algorithm with different smoothing selection strategies as a method for solving the LP relaxation of the MRF energy minimization problem (1).

**Paper Organization.** Section 2 briefly describes basic notions. Section 3 is devoted to estimation of upper (primal) bounds for the dual objective and its smooth approximation. Section 4 describes possible strategies of smoothing selection. And finally Sections 5 and 6 contain results of experimental evaluation and conclusions respectively.

## 2 PRELIMINARIES

We denote by $\mathcal{G} = (\mathcal{V}, \mathcal{E})$ the undirected graph with nodes $\mathcal{V}$ and edges $\mathcal{E} \subset \mathcal{V} \times \mathcal{V}$. Associated with the nodes $v \in \mathcal{V}$ are finite *sets of labels* $\mathcal{X}_v$. The Cartesian product set $\mathcal{X} = \otimes_{v \in \mathcal{V}} \mathcal{X}_v$ is the space of possible *labelings* $x \in \mathcal{X}$, i.e. a labeling is a collection $(x_v \colon v \in \mathcal{V})$ of labels. We use the short-hand notation $x_{uv}$ for a pair of labels $(x_u, x_v)$, and $\mathcal{X}_{uv}$ for the corresponding space $\mathcal{X}_u \times \mathcal{X}_v$. The cost functions $\theta_v \colon \mathcal{X}_v \to \mathbb{R}$, $v \in \mathcal{V}$, and $\theta_{uv} \colon \mathcal{X}_{uv} \to \mathbb{R}$, $uv \in \mathcal{E}$, are referred to as *unary* and *pairwise potentials*, respectively, with the complete set of potentials associated with the graph denoted as $\theta$.

As mentioned in the introduction, our objective function is the dual of a linear programming relaxation of the energy minimization problem (1). As is widely known (see e.g. [24]), the linear programming relaxation of the problem (1) reads

$$\min_\mu \sum_{v \in \mathcal{V}} \sum_{x_v \in \mathcal{X}_v} \theta_{v,x_v} \mu_{v,x_v} + \sum_{uv \in \mathcal{E}} \sum_{x_{uv} \in \mathcal{X}_{uv}} \theta_{uv,x_{uv}} \mu_{uv,x_{uv}}$$

$$\text{s.t.} \begin{cases} \sum_{x_v \in \mathcal{V}} \mu_{v,x_v} = 1, \ v \in \mathcal{V}, \\ \sum_{x_v \in \mathcal{V}} \mu_{uv,x_{uv}} = \mu_{u,x_u}, \ x_u \in \mathcal{X}_u, \ uv \in \mathcal{E}, \\ \sum_{x_u \in \mathcal{V}} \mu_{uv,x_{uv}} = \mu_{v,x_v}, \ x_v \in \mathcal{X}_v, \ uv \in \mathcal{E}, \\ \mu_{uv,x_{uv}} \geq 0, \ x_{uv} \in \mathcal{X}_{uv}, \ uv \in \mathcal{E}. \end{cases} \quad (2)$$

Such a formulation in terms of relaxed indicator vectors $\mu$ is commonly referred to as the overcomplete representation of a discrete graphical model, in which the *local polytope* $\mathcal{L}(\mathcal{G})$, as defined by the constraints of (2), represents a simpler outer bound to the marginal polytope, i.e. the convex hull of all indicator vectors of allowable labelings, cf. [23]. The linear program (2) can be written in short as $\min_{\mu \in \mathcal{L}(\mathcal{G})} \langle \theta, \mu \rangle$. Abusing notation we will denote $\langle \theta, \mu \rangle$ by $E(\mu)$.

It turns out that a dual problem for (2) has a simpler structure and less variables, hence its optimization is easier than the direct optimization of (2). There are several ways how the dual problem can be derived and represented. We will follow *a dual decomposition approach* [21, 9], since it allows for efficient optimization of real-sized problems (see a comparison in [19]). For the sake of simplicity we consider a special case typical for grid graphs.

Let $\mathcal{G}^i = (\mathcal{V}^i, \mathcal{E}^i)$, $i \in \{1, 2\}$, be two *acyclic* subgraphs of the *master graph* $\mathcal{G}$. Let $\mathcal{V}^1 = \mathcal{V}^2 = \mathcal{V}$, $\mathcal{E}^1 \bigcup \mathcal{E}^2 = \mathcal{E}$ and $\mathcal{E}^1 \bigcap \mathcal{E}^2 = \emptyset$ (e.g. if $\mathcal{G}$ is a grid graph, $\mathcal{E}^1$ may contain all horizontal edges of $\mathcal{G}$ and $\mathcal{E}^2$ all vertical ones). Then the overall energy becomes the sum of the energies corresponding to these subgraphs,

$$E_{\mathcal{G}}(\theta, x) = \sum_{i=1}^{2} \sum_{v \in \mathcal{V}^i} \theta^i_{v,x_v} + \sum_{uv \in \mathcal{E}^i} \theta^i_{uv,x_{uv}}$$
$$= E_{\mathcal{G}^1}(\theta^1, x) + E_{\mathcal{G}^2}(\theta^2, x), \quad (3)$$

provided $\theta^i_{uv} = \theta_{uv}$, $uv \in \mathcal{E}^i$, $i = 1, 2$ and $\theta^1_{v,x_v} + \theta^2_{v,x_v} = \theta_{v,x_v}$, $\forall v \in \mathcal{V}, x_v \in \mathcal{X}_v$. The latter condition can be represented in a parametric way as $\theta^1_{v,x_v} = \frac{\theta_{v,x_v}}{2} + \lambda_{v,x_v}$ and $\theta^2_{v,x_v} = \frac{\theta_{v,x_v}}{2} - \lambda_{v,x_v}$, $v \in \mathcal{V}$, $x_v \in \mathcal{X}_v$, where $\lambda_{v,x_v} \in \mathbb{R}$. The dual space $\{\lambda_{v,x_v} \in \mathbb{R} | v \in \mathcal{V}, x_v \in \mathcal{X}_v\}$ will be denoted as $\Lambda$. Thus we consider $\theta^i$ as a function of $\lambda$ and obviously have

$$\min_{x \in \mathcal{X}} E_{\mathcal{G}}(\theta, x) \geq \max_{\lambda \in \Lambda} \sum_{i=1}^{2} \min_{x \in \mathcal{X}} E_{\mathcal{G}^i}(\theta^i(\lambda), x). \quad (4)$$

It has been shown [10] that all decompositions into acyclic subgraphs which cover the original graph yield the same lower bound as the one in (4). Furthermore, it is established that this bound equals the solution of the linear programming problem (2).

Hence, in order to solve the relaxed problem (2) we will maximize its dual, the concave but non-smooth function $U$ defined by the maximand of (4), i.e.

$$U(\lambda) = \sum_{i=1}^{2} \min_{x \in \mathcal{X}} E_{\mathcal{G}^i}(\theta^i(\lambda), x). \quad (5)$$

We note that the function $U(\lambda)$ can be easily evaluated by dynamic programming for any acyclic graph $\mathcal{G}^i$.

**Smoothed Dual Objective and its Primal Counterpart.** There is a number of approaches to maximize (5). Along with [6, 25, 7, 15] we propose to use a smoothing technique [12] to obtain a smooth approximation of $U(\lambda)$. This would allow both to utilize powerful smooth optimization algorithms (e.g. [15]) and to obtain guarantees for convergence to the optimum of the smooth (and as we will see strictly convex) function for efficient (block-)coordinate descent methods similar to TRW-S [8].

The way we construct the smooth approximation is dictated by the need to preserve the efficiency of the decomposition. The smooth function should be easily computable over subgraphs $\mathcal{G}^i$, like its non-smooth counterpart $U$.

Each summand in (5) may be written as the inner product of the potential vector $\theta^i$ and a suitable binary indicator vector $\phi(x)$, i.e.

$$U^i(\lambda) := \min_{x \in \mathcal{X}} E_{\mathcal{G}^i}(\theta^i(\lambda), x) = \min_{x \in \mathcal{X}} \langle \theta^i(\lambda), \phi(x) \rangle. \quad (6)$$

As this minimum is non-smooth in $\lambda$, the same holds for the objective function (5). To obtain a smooth approximation, we replace $\min$ (or rather $-\max$) by the well-known log-sum-exp (or soft-max) function (cf. [12]), yielding

$$\tilde{U}^i_\rho(\lambda) = -\rho \log \sum_{x \in \mathcal{X}} \exp \langle -\theta^i(\lambda)/\rho, \phi(x) \rangle \quad (7)$$

with *smoothing parameter* $\rho$. The resulting function $\tilde{U}^i_\rho$ uniformly approximates $U^i$, that is

$$\tilde{U}^i_\rho(\lambda) + \rho \log |\mathcal{X}| \geq U^i(\lambda) \geq \tilde{U}^i_\rho(\lambda). \quad (8)$$

This directly implies

$$\tilde{U}_\rho(\lambda) + 2\rho \log |\mathcal{X}| \geq U(\lambda) \geq \tilde{U}_\rho(\lambda) \quad (9)$$

for $\tilde{U}_\rho := \sum_{i=1}^{2} \tilde{U}^i_\rho$.

Please note that for acyclic graphs $\mathcal{G}^i$ evaluating $\tilde{U}^i_\rho$ (and thus $\tilde{U}_\rho$) is as easy as $U^i$, and can be done by dynamic programming.

Inequality (9) provides a possibility to exchange optimization of the non-smooth function $U$ with the optimization of its smooth approximation $\tilde{U}_\rho$. Selecting the smoothing parameter $\rho$ small enough (see Section 4.1) or properly decreasing it during optimization (Section 4.2), one can guarantee attainment of the optimum of $U$ with a given precision.

Additionally for $\rho = 1$ the function $\tilde{U}_\rho$ has another important meaning: it is a dual function for *the tree-reweighted free energy* [22] and can be used to estimate approximate marginals [5, 3, 11].

Let us denote by $N_{uv}$ the number of subgraphs containing the edge $uv \in \mathcal{E}$ and by $N_v$ the number of subgraphs containing the node $v \in \mathcal{V}$. In our special case $N_{uv} = 1$ and $N_v = 2$.

**Theorem 1 ([22, 25])** *The tree-reweighted free energy*

$$\tilde{E}_\rho(\mu) := \langle \theta, \mu \rangle - \rho \bigg( \sum_{v \in \mathcal{V}} \sum_{x_v \in \mathcal{X}_v} N_v \mu_{v,x_v} \log \mu_{v,x_v}$$
$$+ \sum_{uv \in \mathcal{E}} \sum_{x_{uv} \in \mathcal{X}_{uv}} N_{uv} \mu_{uv,x_{uv}} \log \frac{\mu_{uv,x_{uv}}}{\mu_{v,x_v} \mu_{u,x_u}} \bigg) \quad (10)$$

*defined over $\mathcal{L}(\mathcal{G})$ is a Lagrange dual for $\tilde{U}_\rho(\lambda)$. And since strict duality holds $\min_{\mu \in \mathcal{L}(\mathcal{G})} \tilde{E}_\rho(\mu) = \max_{\lambda \in \Lambda} \tilde{U}_\rho(\lambda)$.*

**S-TRWS Algorithm** There are a number of approaches to optimize $\tilde{U}_\rho$ (see e.g. [11, 15, 5]). In this paper we evaluate the S-TRWS algorithm originally devoted to computation of the function $\tilde{U}_\rho$ with a fixed value $\rho = 1$ [11]. Contrary to this we will operate $\rho$ in a way to obtain a good (up to a given precision) approximation for the maximum of the non-smooth function $U$. We selected this algorithm because its original, non-smoothed analogue [8] is one of the most efficient schemes for optimizing $U$. The second advantage of S-TRWS is that it is guaranteed to converge to the optimum of $\tilde{U}_\rho$ for any fixed $\rho$ contrary to TRW-S, which does not converge to the optimum of $U$ in general.

We introduce vector of "marginals" $\nu_\rho^i(\lambda) \in \mathbb{R}^{\sum_{w \in \mathcal{V} \cup \mathcal{E}} |\mathcal{X}_w|}$, $i \in \{1, 2\}$ by

$$\nu_\rho^i(\lambda)_{w, x_w} := \frac{\sum_{x' \in \mathcal{X}, x'_w = x_w} \exp \langle -\theta^i(\lambda)/\rho, \phi(x') \rangle}{\exp(-\tilde{U}_\rho^i(\lambda)/\rho)} . \quad (11)$$

It is well-known (see e.g. [15, Lemma 1]), that the coordinates of the gradient $\nabla \tilde{U}_\rho \in \mathbb{R}^{\sum_{v \in \mathcal{V}} |\mathcal{X}_v|}$ are equal to

$$\nabla \tilde{U}_\rho(\lambda)_{v, x_v} = \nu_\rho^1(\lambda)_{v, x_v} - \nu_\rho^2(\lambda)_{v, x_v} . \quad (12)$$

The S-TRWS algorithm is a specially organized coordinate descent procedure applied to the smoothed function $\tilde{U}_\rho$. It sequentially updates variables $\lambda$ according to the rule

$$\lambda_{v, x_v} := \lambda_{v, x_v} + (\log \nu_\rho^1(\lambda)_{v, x_v} - \log \nu_\rho^2(\lambda)_{v, x_v})/2 . \quad (13)$$

Applying this rule corresponds to a coordinate ascent step w.r.t. variable $\lambda_{v, x_v}$. The power of this algorithm stems from an extremely efficient way of processing updates (13). Namely, updates are done sequentially, but the computational cost to perform all these $\sum_{v \in \mathcal{V}} |\mathcal{X}_v|$ updates is basically the same as just computing gradient coordinates $\nu_\rho^i(\lambda)$ (see [8] for details), which makes it superior to any general-purpose gradient based algorithm. Due to (12), it is easy to see that an optimum of $\tilde{U}_\rho$ is a fix-point of this algorithm. We refer to [11] for a proof of convergence. We also note that the implementation of this algorithm is far from trivial because of the necessity to exponentiate very large numbers, corresponding to small values of $\rho$ in (11). We coped with this difficulty efficiently based on the idea provided in [12, p.140].

## 3 UPPER BOUNDS ESTIMATION

There are at least two reasons why it is important to be able to reconstruct a sequence of primal feasible points $\mu^t \in \mathcal{L}(\mathcal{G})$ converging to a solution of the problem (2) or (10), from a sequence $\lambda^t$ converging to the maximum of the corresponding dual objective. The first reason is the values $\mu^t$ themselves: for $\rho = 1$ they can be considered as approximate marginal probabilities [27, 22]. The second reason is an upper bound $E(\mu^t)$ or $\tilde{E}_\rho(\mu^t)$, which can be used for sound stopping criteria [15]. We will use the upper bound to construct a diminishing sequence of smoothing parameters $\rho$ to get an $\varepsilon$-approximation for a maximum of the non-smooth function $U$ via optimization of its smooth approximation $\tilde{U}_\rho$, see Section 4.2.

**Primal LP Objective Estimation** The simplest (and thus quite popular) way of computing upper bounds for maximum of (5) is rounding schemes [14]. It allows to estimate an integer solution $x \in \mathcal{X}$, from a dual iterate $\lambda$. However, the energy $E(\theta, x)$ corresponding to the integer solution $x$ may not achieve the minimal value of objective (2) even in case the labeling was estimated based on the optimal value $\lambda^*$ delivering the maximum of the dual objective $U$.

Alternatively we use the primal LP-bound construction introduced in [15]. We denote by $\mathbb{R}_+(\mathcal{G}) = \mathbb{R}_+^{\sum_{v \in \mathcal{V}} |\mathcal{X}_v| + \sum_{uv \in \mathcal{E}} |\mathcal{X}_{uv}|}$ a nonnegative linear half-space containing the local polytope $\mathcal{L}(\mathcal{G})$. Additionally we use the notation $\mathcal{L}_{uv}(\mu_u, \mu_v) \subset \mathbb{R}_+^{|\mathcal{X}_{uv}|}$ for the domain of pairwise marginals $\mu_{uv}$, $uv \in \mathcal{X}_{uv}$ satisfying the constraints of (2) for given unary marginals $\mu_u, \mu_v$. We use the following reformulation of [15, Thm. 2]:

**Theorem 2** *Let $\hat{\mu}_\rho(\lambda) \in \mathbb{R}_+(\mathcal{G})$ be computed as*

$$\hat{\mu}_\rho(\lambda)_{v, x_v} = \frac{\nu_\rho^1(\lambda)_{v, x_v} + \nu_\rho^2(\lambda)_{v, x_v}}{2} , \quad (14)$$

$$\hat{\mu}_\rho(\lambda)_{uv, x_{uv}} = \underset{\mu_{uv, x_{uv}} \in \mathcal{L}_{uv}(\hat{\mu}_\rho(\lambda)_u, \hat{\mu}_\rho(\lambda)_v)}{\arg \min} \langle \theta_{uv}, \mu_{uv} \rangle . \quad (15)$$

*Then for $\tilde{\lambda} = \arg \max_{\lambda \in \Lambda} \tilde{U}_\rho(\lambda)$ holds*

$$U(\tilde{\lambda}) + 2\rho \log |\mathcal{X}| \geq E(\hat{\mu}_\rho(\tilde{\lambda})) \geq U(\lambda), \ \lambda \in \Lambda . \quad (16)$$

The inequality (16) basically states that the bound $E(\hat{\mu}_\rho(\tilde{\lambda}))$ becomes exact as $\rho$ vanishes.

As noted in [15], evaluating (15) constitutes a *transportation problem* – a well-studied class in linear programing. Since the size of each individual problem is small, these can be easily solved by any appropriate method of linear programming. We found a specialization of the simplex method [1] to be quite efficient in our case.

**Tree-Reweighted Free Energy Estimation.** The upper bound for the minimum of the tree-reweighted energy (10) can be estimated in the same manner as in Theorem 2.

**Theorem 3** *Let $\tilde{\mu}_\rho(\lambda) \in \mathbb{R}_+(\mathcal{G})$ be computed as*

$$\tilde{\mu}_\rho(\lambda)_{v, x_v} = \frac{\nu_\rho^1(\lambda)_{v, x_v} + \nu_\rho^2(\lambda)_{v, x_v}}{2} ,$$

$$\tilde{\mu}_\rho(\lambda)_{uv, x_{uv}} = \underset{\mu_{uv, x_{uv}} \in \mathcal{L}_{uv}(\tilde{\mu}_\rho(\lambda)_u, \tilde{\mu}_\rho(\lambda)_v)}{\arg \min} \langle \theta_{uv}, \mu_{uv} \rangle$$

$$- \rho \sum_{x_{uv} \in \mathcal{X}_{uv}} N_{uv} \mu_{uv, x_{uv}} \log \frac{\mu_{uv, x_{uv}}}{\mu_{v, x_v} \mu_{u, x_u}} . \quad (17)$$

*Then $\tilde{E}_\rho(\tilde{\mu}_\rho(\lambda)) \geq U(\lambda')$ for any $\lambda, \lambda' \in \Lambda$ and for $\tilde{\lambda} = \arg\max_{\lambda \in \Lambda} \tilde{U}_\rho(\lambda)$ holds $U(\tilde{\lambda}) = \tilde{E}_\rho(\tilde{\mu}_\rho(\tilde{\lambda}))$.*

Please note that evaluating (17) constitutes a small-sized convex problem (an entropy maximization kind problem) with linear constraints. It can be efficiently solved e.g. by interior point methods.

As we already mentioned, Theorem 3 defines a method for estimating an upper bound for the optimal value of the tree-reweighted energy (10), which can be utilized to constructing a sound stopping criterion based on the duality gap for algorithms maximizing $\tilde{U}_\rho$.

## 4 SMOOTHING SCHEDULING

In this section we will discuss how to estimate an optimum of $U$ via optimization of its smooth approximation $\tilde{U}_\rho$, i.e. how the smoothing parameter $\rho$ can be selected or/and updated to guarantee the achievement of an $\varepsilon$-approximate solution of $U$.

We consider two approaches of controlling the smoothing gap $\Delta = \max_{\lambda \in \Lambda} U(\lambda) - \tilde{U}_\rho(\lambda)$: precision oriented and diminishing smoothing. Each of them will be considered in connection to both worst-case and adaptive estimation of the smoothing parameter $\rho$ for a given value of $\Delta$.

### 4.1 PRECISION-ORIENTED SMOOTHING

**Precision-Oriented Worst-Case Smoothing Parameter Selection.** The simplest way to select the parameter $\rho$ is to fix it small enough. Let $\lambda^*$ and $\tilde{\lambda}$ denote optimal points of $U$ and $\tilde{U}_\rho$ respectively. By applying the inequality (9) to $\tilde{\lambda}$ and $\lambda^*$ and taking into account that $\tilde{U}_\rho(\tilde{\lambda}) \geq \tilde{U}_\rho(\lambda^*)$ and $U(\lambda^*) \geq U(\tilde{\lambda})$ we obtain

$$\tilde{U}_\rho(\tilde{\lambda}) + \underbrace{2\rho \log |\mathcal{X}|}_{=:\Delta(\rho)} \geq U(\lambda^*) \geq \tilde{U}_\rho(\tilde{\lambda}). \quad (18)$$

Hence if the inequality

$$\Delta(\rho) < \varepsilon \quad (19)$$

holds, the optimal value $U(\lambda^*)$ can be estimated with precision $\varepsilon$ by optimizing $\tilde{U}_\rho$. Inequality (19) can be transformed into a range of allowed values for the smoothing parameter: $0 \leq \rho < \frac{\varepsilon}{2 \log |\mathcal{X}|}$. However selecting an optimal value $\rho$ from this range is not straightforward and depends on the algorithm used for the optimization of $\tilde{U}_\rho$. Lemma 3 in [15] claims that if the convergence rate of the optimization algorithm reads $O(\frac{1}{\rho})$ as a function of $\rho$, then the optimal value of $\rho$ equals to a half of the upper bound of the range, i.e.

$$\rho = \frac{\varepsilon}{4 \log |\mathcal{X}|}. \quad (20)$$

However since little is known about the convergence rate of the considered S-TRWS algorithm, we employ this lemma as a hypothesis.

**Precision-Oriented Adaptive Smoothing Parameter Selection.** Indeed, as mentioned in [15], the estimation $2\rho \log |\mathcal{X}|$ for the smoothing gap is typically loose. Hence it was proposed by the authors to perform a local estimation the smoothing gap as

$$\hat{\Delta}^t(\rho) := \max_t U(\lambda^t) - \tilde{U}_\rho(\lambda^t) \quad (21)$$

for iterates $\lambda^t$ of the optimization algorithm. The value $\rho$ has to be selected such that $\hat{\Delta}^t(\rho) < \varepsilon$, even more, using Lemma 3 in [15] it was proposed to select $\rho$ to satisfy

$$\hat{\Delta}^t(\rho) < \frac{\varepsilon}{2}. \quad (22)$$

Since this simple (and seemingly quite natural!) approach guarantees fulfillment of (19) only locally with respect to the sequence $\{\lambda^t\}$, it does not guarantee attainment of

**(i)** precision $\varepsilon$ in general. Latter can be demonstrated by an example, when the difference $U(\tilde{\lambda}) - \tilde{U}_\rho(\tilde{\lambda})$ in an optimum $\tilde{\lambda}$ of the smoothed function $\tilde{U}_\rho$ is smaller than the one $U(\lambda^*) - \tilde{U}_\rho(\lambda^*)$ in the optimum $\lambda^*$ of the non-smooth one, $U$, and $U(\lambda^*) - \tilde{U}_\rho(\tilde{\lambda}) > \varepsilon$. Starting an algorithm optimizing $\tilde{U}_\rho$ in the point $\tilde{\lambda}$ with

$$\hat{\Delta}^t(\rho) = U(\tilde{\lambda}) - \tilde{U}_\rho(\tilde{\lambda}) < \frac{\varepsilon}{2} \quad (23)$$

makes it stall in the starting point, since (a) the point $\tilde{\lambda}$ is already an optimum of $\tilde{U}_\rho$ and the algorithm can not get better with respect to $\tilde{U}_\rho$; (b) changing of the smoothing $\rho$ is not performed since (22) holds.

**(ii)** the stopping condition $E(\mu_\rho(\lambda^t)) - U(\lambda^t) \leq \varepsilon$ even if (22) holds together with

$$U(\lambda^*) - \tilde{U}_\rho(\tilde{\lambda}) \leq \frac{\varepsilon}{2}. \quad (24)$$

Condition (24) guarantees only the attainment of the accuracy $\varepsilon$ for the dual objective $U$, but not for the duality gap.

Even though we did not encounter such unfortunate cases in practice, there is a general cure for these. Due to Theorem 3 we can estimate the Lagrange dual function $\tilde{E}_\rho$ for $\tilde{U}_\rho$ and decrease $\rho$ by a factor $\eta > 1$ together with restarting the algorithm from the current point as soon as

$$\tilde{E}_\rho(\tilde{\mu}_\rho(\lambda^t)) - \tilde{U}_\rho(\lambda^t) > \frac{\varepsilon}{2} \quad (25)$$

is not satisfied. In other words, one has to select the smoothing parameter $\rho$ such that both conditions (22) and (25) hold. We prove this in the supplementary material because of the space limitations.

**Algorithm 1** Generic diminishing smoothing algorithm.

- Given: target solution accuracy $\varepsilon$
- Initialize: $\lambda^0 \in \Lambda$, $\rho^0 > 0$, $E_{\min}^0 = E(\hat{\mu}_{\rho^0}(\lambda^0))$.
- Iterate (for $t \geq 0$):
  1. **If** $E_{\min}^t - U(\lambda^t) < \varepsilon$  **Exit and return** $\lambda^t$.
  2. $\rho^{t+1} := \min\left\{\rho^t, F(E_{\min}^t, \tilde{U}_{\rho^t}(\lambda^t))\right\}$.
  3. $\lambda^{t+1} := \psi[\tilde{U}_{\rho^{t+1}}](\lambda^t)$.
  4. $E_{\min}^{t+1} := \min\{E_{\min}^t, E(\hat{\mu}_{\rho^{t+1}}(\lambda^{t+1}))\}$.
  5. $t := t+1$.  **Goto** 1.

## 4.2 DIMINISHING SMOOTHING

Instead of using a fixed value of $\rho$ satisfying (19) or keeping it rarely changed when estimation $\hat{\Delta}(\rho)$ of the smoothing gap does not satisfy (22), we will construct a diminishing $\rho$-sequence depending on the duality gap. This idea is based on the assumption that maximization of $\tilde{U}_\rho$ is more efficient for larger values of $\rho$.

Let us consider a generic algorithm - Algorithm 1 - for the diminishing smoothing. Step 1 checks the stopping condition based on the duality gap. Minimization in steps 2 and 4 is required to ensure monotone decreasing of $\rho$ and to keep the best upper bound $E_{\min}^t$ so far. A (for the moment) unknown mapping $F\colon \mathbb{R} \times \mathbb{R} \to \mathbb{R}_+$ determines how the smoothing parameter $\rho$ should be updated. The mapping $\psi[\tilde{U}_\rho]$ corresponds to one or more iterates of the S-TRWS algorithm (see (11) and (13)) applied to the function $\tilde{U}_\rho$.

Our goal is to determine such mappings $F$, which at least guarantee convergence of Algorithm 1 iterates to the optimum of the non-smooth function $U$.

**Definition 1** *A mapping* $\psi[\tilde{U}.]\colon \mathbb{R}_+ \times \Lambda \to \Lambda$ *is called an optimizing mapping if*
*(i) for any $\rho > 0$ and any $\lambda^0 \in \Lambda$ the iterative process $\lambda^{t+1} = \psi[\tilde{U}_\rho](\lambda^t)$ converges to the set $\Lambda^*(\tilde{U}_\rho)$ of optimal solutions of $\tilde{U}_\rho$;*
*(ii) $\tilde{U}_\rho(\psi[\tilde{U}_\rho](\lambda)) \geq \tilde{U}_\rho(\lambda)$, moreover for $\lambda \notin \Lambda^*(\tilde{U}_\rho)$ holds $\tilde{U}_\rho(\psi[\tilde{U}_\rho](\lambda)) > \tilde{U}_\rho(\lambda)$,*
*(iii) $\psi[\tilde{U}_\rho](\lambda)$ is continuous w.r.t. $\rho$ for any $\lambda \in \Lambda$.*

Due to continuity of $\exp, \sum$ and $\log$, the iterates of the S-TRWS algorithm depend continuously on $\rho$. Theorem 6 in [25] proves that every iteration of the algorithm strictly increases the value of $\tilde{U}_\rho$ for a fixed $\rho$; the convergence of the S-TRWS algorithm to the optimum of the function $\tilde{U}_\rho$ for any fixed $\rho$ is proved in [11]. Hence one or more iterations of the algorithm satisfy the conditions of Def. 1. We explain the practical importance of using more than a single iteration of the algorithm at the end of this paragraph.

**Theorem 4** *Let* $\psi[\tilde{U}.]\colon \mathbb{R}_+ \times \Lambda \to \Lambda$ *be an optimizing mapping and let the mapping* $F\colon \mathbb{R} \times \mathbb{R} \to \mathbb{R}_+$ *satisfy the condition*

$$\Delta\left(F(E_{\min}^t, \tilde{U}_{\rho^t}(\lambda^t))\right) \leq \frac{E_{\min}^t - \tilde{U}_{\rho^t}(\lambda^t)}{2} \qquad (26)$$

*where equality holds only when $E_{\min}^t - \tilde{U}_{\rho^t}(\lambda^t) = 0$. Then iterates $\lambda^t$ of Algorithm 1 converge to the optimum of the function $U$.*

**Proof.** From the first property of an optimizing mapping (see Def. 1) follows that it suffices to prove that $\rho^t \xrightarrow{t\to\infty} 0$ and $\rho^t > 0$, $\forall t$. Since $E(\mu) > \tilde{U}_\rho(\lambda)$ for any $\mu, \lambda \notin \Lambda^*(U)$ and $\rho \geq 0$, from step 2 follows that $\rho^t > 0$, $\forall t$. Sequence $\rho^t$ is monotone and bounded by 0 from below, thus it converges to some $\rho^* \geq 0$. We will prove that $\rho^* = 0$. Let $\rho^* \neq 0$. From (26), which is strict while $E_{\min}^t - \tilde{U}_{\rho^t}(\lambda^t) \neq 0$, follows that $\tilde{U}_{\rho^t}(\lambda^t) + \Delta(\rho^{t+1}) < E(\hat{\mu}_\rho(\lambda^t)) - \Delta(\rho^{t+1})$. Taking the limit as $t$ tends to infinity, denoting $\tilde{\lambda} = \arg\min_{\lambda \in \Lambda} \tilde{U}_{\rho^*}(\lambda)$ and using continuity in $\rho$ and $\lambda$ of $U(\lambda), \tilde{U}_\rho(\lambda), E(\hat{\mu}_\rho(\lambda))$ and $\psi$ we obtain $\tilde{U}_{\rho^*}(\tilde{\lambda}) + \Delta(\rho^*) < E(\hat{\mu}_{\rho^*}(\tilde{\lambda})) - \Delta(\rho^*)$. Since $\Delta(\rho) = 2\log|\mathcal{X}|$, from (16) follows $E(\hat{\mu}_{\rho^*}(\tilde{\lambda})) - \Delta(\rho^*) \leq U(\tilde{\lambda})$ and thus $\tilde{U}_{\rho^*}(\tilde{\lambda}) + \Delta(\rho^*) < U(\tilde{\lambda})$. From (9) follows the contradiction, which implies $\rho^* = 0$. □

As follows from the proof nothing prevents to improve the upper bound $E_{\min}^t$ by using any available method of its estimation align with the one provided by Theorem 2. In our experiments we use rounding schemes [14], which sometimes allows to get better estimation of $E_{\min}^t$ and hence speed-up Algorithm 1.

Number of cycles of the S-TRWS algorithm, used for computing the optimizing mapping $\psi$, should be carefully selected because of the fact, that performing 1 cycle of the S-TRWS without prior change of the smoothing parameter is computationally 2 times cheaper than performing it with the change. This is caused by peculiarities of the S-TRWS algorithm. Namely, to compute values $\nu_\rho^i(\lambda)_{v,x_v}$ (see (11)) it has to perform both so-called *forward* and *backward* moves of a dynamic programming algorithm. And if after a given run no changes where made to smoothing, the result of the forward (backward) move do not need to be recomputed. It means, that computing $n$ cycles of the S-TRWS algorithm with a *constant* smoothing costs only $n + 1$ oracle calls (additional one oracle call for the initial forward move) instead of $2n$.

**Diminishing Worst-Case Smoothing Parameter Selection.** Theorem 4 implies that for some $\gamma > 1$

$$F(E_{\min}^t, \tilde{U}_{\rho^t}(\lambda^t)) = \Delta^{-1}\Big(\frac{E_{\min}^t - \tilde{U}_{\rho^t}(\lambda^t)}{2\gamma}\Big), \qquad (27)$$

at the step 2 of Algorithm 1. Here $\gamma$ is some constant (which indeed can differ from iteration to iteration)

and $\Delta^{-1}(\cdot)$ denotes the inverse mapping to $\Delta(\cdot)$. Since $\Delta(\rho) = 2\rho \log |\mathcal{X}|$ it reads $\Delta^{-1}(A) = \frac{A}{2 \log |\mathcal{X}|}$. Hence

$$F(E_{\min}^t, \tilde{U}_{\rho^t}(\lambda^t)) = \frac{E_{\min}^t - \tilde{U}_{\rho^t}(\lambda^t)}{4\gamma \log |\mathcal{X}|}, \gamma > 1, \quad (28)$$

and Algorithm 1 is defined up to a parameter $\gamma$. We refer to Section 5 for experimental evaluation of this parameter.

**Diminishing Adaptive Smoothing Parameter Selection.** As we already mentioned the estimate $2\rho \log |\mathcal{X}|$ for the smoothing gap is often too loose in practice, which can potentially slow down optimization algorithms. In this section we show how the adaptive estimate (21) can be used together with an appropriate approximation of the inverse mapping $\Delta^{-1}$ to speed up the algorithm. Additionally we will use a procedure similar to the one presented in (25) to prevent the algorithm from stalling.

The method is based on a local approximation of $\hat{\Delta}^t(\rho)$ near the current point $\lambda$ with an affine function $\hat{\Delta}^t(\rho) := \delta \cdot \rho + \alpha$, where $\delta = \frac{\partial \hat{\Delta}^t}{\partial \rho} = \frac{\partial (U - \tilde{U}_\rho)}{\partial \rho} = -\frac{\partial \tilde{U}_\rho}{\partial \rho}$ and assigning $\hat{\Delta}^{-1}(d) := (d - \alpha)/\delta$. Hence step 2 of Algorithm 1 takes the form

$$\delta^t := -\frac{\partial \tilde{U}_{\rho^t}}{\partial \rho}(\lambda^t), \quad \alpha^t := \hat{\Delta}^t - \delta^t \rho^t, \quad (29)$$

$$\rho^{t+1} := \min \left\{ \rho^t, \left( \frac{E_{\min}^t - \tilde{U}_{\rho^t}(\lambda^t)}{2\delta^t \gamma} - \frac{\alpha^t}{\delta^t} \right) \right\}. \quad (30)$$

Parameter $\alpha^t$ of this approximation depends on the estimate $\hat{\Delta}^t$ defined by (21).

To avoid stalling of the algorithm one should decrease $\rho$ as soon as

$$\tilde{E}_\rho(\mu_\rho(\lambda^t)) - \tilde{U}_{\rho^t}(\lambda^t) \le \frac{E_{\min}^t - \tilde{U}_{\rho^t}(\lambda^t)}{2\gamma}. \quad (31)$$

The proof for that is based on comparing (23) and (25) to (27) and (31). Note that they become equivalent when $\varepsilon = (E_{\min}^t - \tilde{U}_{\rho^t}(\lambda^t))/\gamma$.

We sum up the above considerations in Algorithm 2. This algorithm additionally depends on the parameter $\eta$ defining how much the smoothing parameter $\rho$ should be decreased. In our experiments we use $\eta = 2$, but we observed the inequality (31) being fulfilled only for $\gamma$ close to 1 (which we found to be non-optimal) and thus the value $\eta$ does not influence results of our experiments.

It remains only to note that the derivative $\frac{\partial \tilde{U}_{\rho^t}(\lambda^t)}{\partial \rho}$ can be computed efficiently, since it is given by a ready-to-apply formula in the following proposition.

**Proposition 1** *Let $\nu_\rho^i(\lambda) \in \mathbb{R}^{\sum_{w \in \mathcal{V} \cup \mathcal{E}} |\mathcal{X}_w|}$ be defined*

**Algorithm 2** Adaptive diminishing smoothing (ADS*al*)

- Given: a prescribed solution accuracy $\varepsilon$.
- Initialize: $\lambda^0 \in \Lambda$, $\rho^0 > 0$, $\gamma > 1$, $\eta > 1$, $E_{\min}^0 = E(\hat{\mu}_{\rho^0}(\lambda^0))$.
- Iterate (for $t \ge 0$):
  1. **If** $E_{\min}^t - U(\lambda^t) < \varepsilon$ **Exit and return** $\lambda^t$.
  2. $\delta^t := -\frac{\partial \tilde{U}_{\rho^t}}{\partial \rho}(\lambda^t), \quad \alpha^t := \hat{\Delta}^t - \delta^t \rho^t$.
  3. $\rho^{t+1} := \min \left\{ \rho^t, \left( \frac{E_{\min}^t - \tilde{U}_{\rho^t}(\lambda^t)}{2\delta^t \gamma} - \frac{\alpha^t}{\delta^t} \right) \right\}$.
  4. **If** $\tilde{E}_\rho(\tilde{\mu}_\rho(\lambda^t)) - \tilde{U}_\rho(\lambda^t) \le \frac{E_{\min}^t - \tilde{U}_{\rho^t}(\lambda^t)}{2\gamma}$ **then** $\rho^{t+1} := \frac{\rho^{t+1}}{\eta}$.
  5. $\lambda^{t+1} := \psi[\tilde{U}_{\rho^{t+1}}](\lambda^t)$.
  6. $E_{\min}^{t+1} := \min\{E_{\min}^t, E(\hat{\mu}_{\rho^{t+1}} \lambda^{t+1})\}$.
  7. $t := t + 1$. **Goto** 1.

*by (11). Then*

$$\frac{\partial \tilde{U}_\rho(\lambda)}{\partial \rho} = \frac{\tilde{U}_\rho(\lambda)}{\rho} - \frac{1}{\rho} \sum_{i=1}^{2} \sum_{w \in \mathcal{V}^i \cup \mathcal{E}^i} \sum_{x_w \in \mathcal{X}_w} \theta^i(\lambda)_{w,x_w} \nu_\rho^i(\lambda)_{w,x_w}.$$

## 5 EXPERIMENTAL EVALUATION

For an experimental evaluation we used datasets from the Middlebury MRF benchmark [20] (datasets *tsukuba, venus, family*) and a computer generated grid model of size $256 \times 256$ with 4 labels. The unary and pairwise factors of this model were randomly selected uniformly in the interval $[0, 1]$. This dataset will be denoted as *artificial*. Complete results of all our experiments are present in the supplementary material. In the main body of the paper we present only subset of plots and a summary Table 5 to save space.

We consider the following four smoothing-based algorithms described in the paper: (i) adaptive diminishing smoothing algorithm (**A-DSal**) – Algorithm 2; (ii) diminishing worst-case smoothing (**WC-DSal**) is defined by Algorithm 1, where the mapping $F$ is specified by (28); (iii) precision oriented adaptive smoothing (**A-STRWS**), where the smoothing degree is controlled by (21), (22) and (25); (iv) precision oriented worst-case smoothing (**WC-STRWS**), where the smoothing degree is fixed and given by (20). In all cases we use the S-TRWS algorithm as an optimizing mapping.

We compare our approaches also to the original **TRW-S** code [8] and to the accelerated first-order Nesterov optimization scheme [15] (**NEST**), for which an implementation was kindly provided by the authors.

Additionally to the upper $\Delta(\rho)$-bound computed due to

Table 1: Results of algorithms evaluation. *Req. OC 0.1%* and *Req. OC 1* – number of oracle calls required to achieve the relative precision 0.1% and the absolute precision 1 respectively; *primal/dual bound*–the best upper/lower bounds achieved during the iterations: about 8000-10000 oracle calls for NEST, TRW-S and other methods. In case the best primal bound is an integer one we print it without decimal point. Please note that A-DSal turned out to be the most efficient solver for all datasets except *family* We did not mark "dual bounds winner" for *tsukuba* dataset, because in this case all algorithms except TRW-S were terminated as soon as the duality gap dropped below 1.

|  | Algorithm | A-DSal | WC-DSal | A-STRWS | WC-STRWS | NEST | TRWS |
|---|---|---|---|---|---|---|---|
| tsukuba | req. OC 0.1% | 52 | 47 | 45 | 33 | 576 | 266 |
| | req. OC 1 | **107** | 151 | 405 | 789 | 6244 | >10000 |
| | primal bound | **369218** | 369218 | 369218 | 369218 | 369218 | 369252 |
| | dual bound | 369217.38 | 369217.12 | 369217.99 | 369217.99 | 369217.99 | 369217.58 |
| venus | req. OC 0.1% | **111** | 131 | 123 | 129 | 1746 | 266 |
| | primal bound | **3047993.47** | 3048546.82 | 3048411 | 3048534.23 | 3050376 | 3048098 |
| | dual bound | **3047965.20** | 3047920.27 | 3047936.20 | 3047934.25 | 3047938.84 | 3047929.95 |
| family | req. OC 0.1% | >8516 | >10006 | >8013 | >8001 | >9023 | **3012** |
| | primal bound | 186636 | 184927 | 185144.02 | 185142.87 | 6136365 | **184825** |
| | dual bound | 184769.13 | 184742.11 | 184735.80 | 184734.96 | 145396.26 | **184788.00** |
| artificial | req. OC 0.1% | **514** | >10004 | 639 | >8001 | 1515 | >10000 |
| | primal bound | **56785.86** | 56838.03 | 56956.08 | 57258.31 | 56815.01 | 81118 |
| | dual bound | 56779.98 | 56777.10 | 56764.87 | 56724.52 | **56780.24** | 56720.56 |

Theorem 2 we also estimate a so-called *integer* upper bound, which corresponds to integer values of variables $\mu$ in (2). The latter bound turns to be often better at the beginning of the optimization process, but contrary to the non-integer one it does not posses a crucial property of convergence to the optimum of the relaxed problem (2).

Since the performance of the precision-oriented smoothing algorithms (A-STRWS, WC-STRWS and NEST) depends on the target precision we run each algorithm on each dataset twice: with target *relative* precision 0.1% and with target *absolute* precision 1. Since potentials of all considered problem from the Middlebury benchmark are integers, getting an integer upper bound within the latter precision would mean solving the corresponding problem exactly in its original, non-relaxed formulation (1). Indeed it has happen for the *tsukuba* dataset – see below.

We use the notion of *oracle calls* for measuring the speed of the algorithms to eliminate the influence of different implementations. One oracle call corresponds to a single (either forward or backward) move of the S-TRWS algorithm. It requires approximately 1.5 seconds for the *tsukuba* dataset, 3.5 s for *venus*, 4.5 s for the *family* and less than 0.5 s for the *artificial* dataset, on a 2.5GHz machine. Such an oracle call approximately corresponds to the oracle call used in NEST, it is 5 times slower than a single (either forward or backward) move of the original non-smooth TRWS algorithm in our implementation and 10 times slower than the original implementation by Kolmogorov [8] for general-form pairwise potentials. The non-smooth TRWS is faster because of the evaluation of the exp-operation needed for smoothing. Preliminary experiments show that simply switching to GPU would already reduce this operation's cost to that of a simple multiplication operation. Hence we ignore the mentioned time difference and consider a single move of the TRW-S algorithm as an oracle call. Additionally we count computations needed to estimate a primal solution $\hat{\mu}_\rho(\lambda^t)$ according to (14)-(15) as an oracle call, since the time it requires is close to that of a single oracle call.

For different values of $\gamma$ and different numbers of inner S-TRWS iterations we measured the number of oracle calls of the A-DSal and WC-DSal algorithms to attain precision 1%. We found $\gamma = 4$ and 3 inner iterations (4 oracle calls) in the optimizing mapping $\psi$ to be optimal for both A-DSal and WC-DSal. Indeed we found the algorithms quite insensitive to the value of $\gamma$, unlike for instance the step-size selection in subgradient-based schemes. Our experiments show similar performance for the interval $2 \leq \gamma \leq 5$.

**Tsukuba dataset. Fig. 1(bottom left)** This dataset seems to be the easiest among investigated. The relative precision 0.1% was achieved by all algorithms within few dozens of oracle calls. Moreover, all algorithms except TRW-S attained the optimum of the non-relaxed problem (1). Number of oracle calls for that, i.e. to get the duality gap less than 1 is the smallest for A-DSal. Other algorithms, described in the paper perform not much worse and indeed much faster than NEST.

**Venus dataset. Fig. 1 (top left)** Contrary to *tsukuba* there remains a big ($\gg 1$) relaxation gap between obtained integer solutions and the lower (dual) bound. Simultaneously the upper (primal) bound for the relaxed LP problem demonstrates a fast convergence to the lower one, which makes A-DSal quite efficient. Other algorithms described in the paper show significantly worse performance. TRW-S is a bit slower at the beginning, but at the end shows the second good result after A-DSal in terms of both upper and lower bounds.

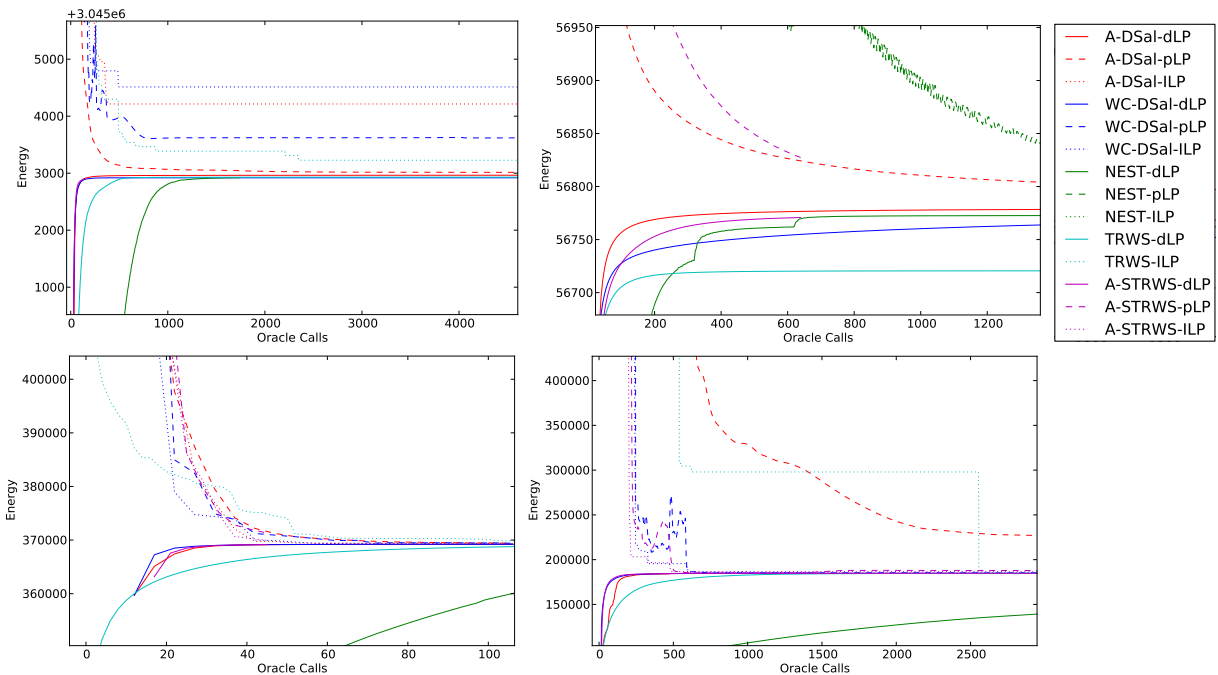

Figure 1: *Venus* (top left), *artificial* (top right), *tsukuba* (bottom left) and *family* (bottom right) datasets. Precision-oriented smoothing algorithms A-STRWS, WC_STRWS and NEST were run with a target precision $0.1\%$. The curves show primal (upper) (*pLP*) and dual (lower) (*dLP*) LP bounds, as well as the best integer primal bound (*ILP*) achieved. Colors correspond to different algorithms – see legend. For all datasets except *tsukua* the A-DSal demonstrates one of the best convergence rates. For the *family* the primal LP bound of A-DSal shows a very slow convergence, which makes the algorithm significantly less efficient.

**Family dataset. Fig. 1(bottom right)** This dataset seems to be quite difficult for dual methods (methods considered in the paper and NEST, TRW-S) because the pairwise factors contain infinite (very large) numbers. This makes dual variables $\lambda$ badly conditioned, i.e. a small change in their values corresponds to a big change of the estimated primal objective. This is why the estimations of the upper (primal) bound constructed according to Theorem 2 converge very slow and thus the smoothing parameter, selected according to the duality gap estimation, takes too large values. This influences A-DSal at most because it uses additionally adaptive smoothing parameter selection, which makes $\rho$ even larger. However even for this setting the lower bound attained by A-DSal outperforms lower bound of all other approaches except TRW-S (see Table 5), which does not suffer from bad primal estimates.

**Artificial dataset. Fig. 1 (top right)** This dataset targets the situations when there is no local evidence in the data. This could be a typical setting when the inference is used in a loop of a learning algorithm, e.g. the structural SVM [2]. At the beginning of the learning process the potentials to be learned could take nearly random values, which makes the problem harder. This dataset clearly shows advantage of methods with an adaptive selection of the smoothing parameter: only those methods attained the precision $0.1\%$. The A-SDal again shows the best performance, whereas TRW-S – the worst one.

**Summary.** As our experiments show, the proposed adaptive diminishing smoothing algorithm (A-DSal) is competitive to or even outperforms the state-of-the-art methods, except in the case when the pairwise factors contain very large values. This issue has to be considered in future work.

## 6 CONCLUSIONS

We proposed an adaptive diminishing smoothing optimization algorithm for an LP relaxation of the energy minimization problem. The algorithm enjoys provable and fast convergence to the optimum, and often outperforms even one of the fastest state-of-the-art methods – TRWS, but contrary to that does never get stuck in non-optimal points.

**Acknowledgement.** This work has been supported by the German Research Foundation (DFG) within the program Spatio-/Temporal Graphical Models and Applications in Image Analysis, grant GRK 1653.